\documentclass[runningheads]{llncs} 

\usepackage{times} \usepackage{xcolor} \usepackage{soul} 
\usepackage[utf8]{inputenc} \usepackage[small]{caption}

\usepackage{latexsym,amsmath,amstext,amsfonts,amsbsy,amssymb,graphicx,hyperref}
\usepackage{times} 

\title{The Foundations of Deep Learning with a Path Towards General Intelligence}

\author{Eray \"Ozkural}
\institute{Celestial Intellect Cybernetics\\celestialintellect.com}

\begin{document}

\maketitle

\begin{abstract} Like any field of empirical science, AI may be approached axiomatically. We formulate requirements for a general-purpose, human-level AI system in terms of postulates. We review the methodology of deep learning, examining the explicit and tacit assumptions in deep learning research. Deep Learning methodology seeks to overcome limitations in traditional machine learning research as it combines facets of model richness, generality, and practical applicability. The methodology so far has produced outstanding results due to a productive synergy of function approximation, under plausible assumptions of irreducibility and the efficiency of back-propagation family of algorithms. We examine these winning traits of deep learning, and also observe the various known failure modes of deep learning. We conclude by giving recommendations on how to extend deep learning methodology to cover the postulates of general-purpose AI including modularity, and cognitive architecture. We also relate deep learning to advances in theoretical neuroscience research.
\end{abstract}

\vspace*{-0.5cm}

\section{Introduction}

\vspace*{-0.3cm}


Deep learning is a rapidly developing branch of machine learning which is clustered around training deep neural models with many layers and rich computational structure well suited to the problem domain \cite{Goodfellow-et-al-2016,juergen-dl}. Initially motivated by modelling the visual cortex \cite{fukushima:1980,Fukushima:2013}, human-level perceptual performance was approached and eventually attained in a number of challenging visual perception tasks such as image recognition with the aid of GPU acceleration \cite{LeCun:89,ranzato-cvpr-07,Graves:09tpami}.
The applications quickly extended to other computer vision tasks such as image segmentation \cite{ciresan2012isbi}, producing a variety of impressive results in visual information processing such as style transfer \cite{image-style-transfer}, opening new vistas in machine learning capabilities. The applications have been extended to domains beyond vision, such as speech recognition \cite{graves:2013icassp}, language processing \cite{kim16-nlp}, and reinforcement learning 
\cite{koutnik:fdg13}
, often with striking performance, proving the versatility and the significance of the approach in AI, urging us to consider whether the approach may yield a general AI (called Artificial General Intelligence (AGI) in some circles), and if so which problems would have to be tackled to make deep learning approach truly human-level AI that covers all aspects of cognition.

We analyze the approach from a 10,000 feet vantage point, revisiting the idea of AI axiomatization. Although, we are generally in agreement with Minsky that the attempt to make AI like physics is likely a futile pursuit, we also note the achievements of later theorists who have applied Bayesian methods successfully. We make no attempt to formalize any of our claims due to space consideration, however we discuss relevant research in cognitive sciences. Then, we apply the same foundational thinking to deep learning critically probing its intellectual foundations. The axioms, or postulates, of AI, are examined with an eye towards whether the current progress in deep learning in some way satisfies them, and what has to be done to fill the gap. The present paper may thus be regarded as an analytical, critical meta-level review, rather than a comprehensive review such as \cite{juergen-dl}.

\vspace*{-0.4cm}

\section{Postulates of General AI}

\vspace*{-0.2cm}

One of the most ambitious mathematical models in AGI research is AIXI \cite{hutter-aixigentle} which is a universal Reinforcement Learning (RL) model that can be applied to a very large variety of AI agent models and AI tasks including game playing, machine learning tasks, and general problem solving. AIXI is based on an extension of Solomonoff's sequence induction model 
which works with arbitrary loss and alphabet \cite{Hutter:03optisp+}, making the aforementioned induction problem fairly general. Hutter proves in his book \cite{uai} that many problems can be easily transformed to this particular formulation of universal induction. There are a few conditions that have to be satisfied for a system to be called a universal induction system, and even then the system must be realized in a practical manner so as to be widely applicable and reproduce the cognitive competencies of homo sapiens, or failing that, a less intelligent animal.

The AIXI model combines Bellman equation with universal induction, casting action selection as the problem of maximizing expected cumulative reward in any computable environment. Although RL is a common approach in machine learning, AIXI had the novelty that it focused solely on universal RL agents. When viewed this way, it is obvious that AIXI is a minimalist cognitive architecture model, that exploits the predictive power of induction in RL setting, that does give the model the kind of versatility noted above. Solomonoff induction presents a desirable limit of inductive inference systems, since it has the least generalization error possible; the error is dependent only on the stochastic source and a good approximation can learn from very few examples \cite{solcomplexity}. AIXI model also retains a property of optimal behavior, Hutter deliberates that the model defines optimal, but incomputable intelligence, and thus any RL agent must approximate it. Therefore, our axiomatization must consider the conditions for Solomonoff's universal  induction model, and consequently AIXI, to be approximated well, but we believe additional conditions are necessary for it to also satisfy generality in practice and within a versatile system, as follows.
\textbf{Completenes:} The class of models that can be acquired by the machine learning system must be Turing-complete. If a large portion of the space of programs is unavailable to the system, it will not have the full power and generalization properties of Solomonoff induction. The convergence theorem in that case is voided, and the generalization performance of Solomonoff induction cannot be guaranteed \cite{solcomplexity}.
\textbf{Stochastic Models:} The system requires an adequately wide class of stochastic models to deal with uncertainty in the real world, a system with only deterministic components will be brittle. Induction is better suited to working with stochastic models, one example of such an approach is Wallace's Minimum Message Length (MML) model where we minimize the message length that contains both the length of the statistical model encoding and data encoding length relative to model \cite{WallaceBoulton:1968,wallace99}. 
\textbf{Bayesian Prediction:} The system must compute the inferences with Bayes' law. The inference in Solomonoff's model is considered Bayesian. In neuroscience, the Bayesian Brain Hypothesis has been mostly accepted, and the brain is often regarded as a Bayesian inference machine that extracts information from the environment in theoretical neuroscience. Jaynes introduced the possibility of Bayesian reasoning in the brain from a statistical point of view \cite{jaynes-reasoning}. The Bayesian approach to theoretical neuroscience is examined in a relatively recent book \cite{bayesianbrain:2007}. Fahlman et. al introduced the statistically motivated energy minimizing Boltzmann machine model \cite{Fahlman:1983}; Hinton et. al connected the induction principle of Minimum Description Length and Helmholtz free energy introducing the autoencoder model in 1993 \cite{Hinton:1993}. Bialek's lab has greatly contributed to the understanding of the Bayesian nature of the brain, a decent summary of the approach detailing the application of the information bottleneck method may be found in \cite{bialek-predictability}. Friston has later rigorously applied the free energy principle and has obtained even more encouraging results, he explains the Bayesian paradigm in \cite{friston-bayesianbrain}.
Note that Helmoltz free energy and the free energy principle are related, and both are related to approximate Bayesian inference. 
\textbf{Principle of Induction:} The system must have a sound principle of induction that is equivalent to Solomonoff's model of induction which uses an a priori probability model of programs that is inversely and exponentially proportional to program size. Without the proper principle of induction, generalization error will suffer greatly, as the system will be corrupted. Likewise, as Solomonoff induction is more completely approximated, the generalization error will decrease dramatically, allowing the system to obtain one-shot learning first predicted by Solomonoff, achieving a successful generalization from a sufficiently complex single example without any prior training whenever such an example is possible.
\textbf{Practical Approximation:} Solomonoff induction has an exponential worst-case bound with respect to program size rendering it infeasible. This surely is not a practical result, any approximation must introduce algorithmic methods to obtain a feasible approximation of the theoretical inductive inference model.
\textbf{Incremental Learning:} The system must be capable of cumulative learning, and therefore it must have a model of memory with adequate practical algorithms. Solomonoff has himself described a rather elaborate approach to transfer learning \cite{solomonoff-incremental}, however, it was not until much later that experimental results were possible for universal induction since Solomonoff's theoretical description did not specify an efficient algorithm. The first such result was obtained in OOPS system \cite{oops} demonstrating significant speedups for a universal problem solver.
\textbf{Modularity and Scalability:} The system must be composed of parametrized modules that attend to different tasks, allowing complex ensemble systems to be built for scalability like the neocortex in the human brain. A monolithic system is not likely to scale well, the system must be able to adapt modules to distinct tasks, and then be able to re-use the skills. A modular system also provides a good base for specialization according to modality and cognitive task, starting from a common module description. In the human brain, there are both functional regions and a complex, hierarchical modular structure in the form of cortical columns, and micro-columns.
\textbf{Cognitive Architecture:} The system must have a cognitive architecture, depending on modularity that will address typical cognitive functions of learning, memory, perception, reasoning, planning, and language as well as aspects of robotics which allow it to control robotic appendages. This manner of organization is modeled after the human brain, however, it seems essential for any real-world AI system that requires these basic competencies to deliver robust performance across a sufficiently general set of cognitive tasks. Even if unlike the brain, the system must have an architectural design, or one that is capable of introducing the required architecture.

These reasonable and desirable properties of a complete AI system lead naturally to a top-down design sometimes called an AGI Unification Architecture among practitioners, if built around the floor plan of a universal induction system such as AIXI. An example of such an approach to designing a cognitive architecture may be seen in \cite{potapov-ga-ppl}. However, this is not necessarily the only kind of solution. An adequate architecture could also be built around a deep learning approach; let us therefore proceed to its postulates.

\vspace*{-0.4cm}

\section{Postulates of Deep Learning}

\vspace*{-0.2cm}

Deep Learning is a particular kind of Artificial Neural Network (ANN) research which shares some commonalities and inherits some assumptions / principles from earlier ANN research some of which may seem implicit to outsiders. We try to recover these tacit or implicit assumptions for the sake of general AI readership, and also delineate the borders of deep learning from other ANN research in the following. \textbf{No Free Lunch:} The well-known No Free Lunch theorem for machine learning 
implies that there can be no general learning algorithm that will be effective for all problems. This theorem has generated a strong bias towards model-based learning in ANN research where the researcher tries to design a rich network model that covers all contingencies in the domain but uses insights into the problem domain and thus the experiment does not suffer from the unreasonable large search space of a model-free learning method. From image processing to language, this particular blend of specificity and generality seems to have guided deep learning quite successfully and resulted in impressive outcomes. The specificity determined by the ANN researcher may be likened to “innateness” in cognitive science. Note that AGI theorists have argued otherwise \cite{freelunch}, therefore this heuristic principle remains arguable.
\textbf{Epistemic Non-reductionism:} This is the view that loosely depends on Quine's observation that epistemic reductionism often fails in terms of explanatory power for the real world \cite{quine-twodogmas}, which is to say that there is a wealth of necessary complexity to account for it. When we look at a deep learning vision architecture, we see that the irreducible patterns of visual information are indeed stored as they are useful however not overmuch; the system does not store every pattern much like our brains. Epistemic irreducibility is a guiding principle in deep learning research, and it is why deep learning models are large rather than small and minimalistic as in some ANN research.
\textbf{Eliminative Materialism:} Churchland's philosophical observation that the brain does not deal in any of the folk psychological concepts in cognitive science literature, but must be understood as the activation state and trajectory of the brain \cite{Churchland1981}, plays a fundamental intellectual role in the deep learning approach, where we shift our attention to brain-like representations and learning for dealing with any problem, even if it looks like a matter of propositional logic to us.
\textbf{Subsymbolic \& Distributed Representation:} Expressed in detail in the classical connectionist volume \cite{Rumelhart:1986:PDP}, this principle is the view that all representations in the brain have a distributed, real-valued representation rather than discrete, symbolic representations that computer scientists prefer in their programs. Sparse Coding hypothesis has been mostly confirmed in neuroscience, therefore we do know that the brain uses population codes that are sparse, distributed, and redundant. Unlike a symbolic representation, the brain networks are fault-tolerant and redundant, and deal with uncertainty at every level. Subsymbolic representations are more robust and better suited to the nature of sensory input. However, we also know that “grandmother cells” exist which may correspond to predicates, which are still best modeled as non-linear detectors, or ReLu units, in a neural network.
\textbf{Universal Approximation:} The universal approximation theorem \cite{Hornik:1991} for multi-layer feed forward neural networks underlies the heuristic of using many hidden layers in a deep learning architecture. The theorem shows that a multi-layer neural network can approximate arbitrary continuous real-valued functions. Therefore, the system is capable of representing any mapping under mild assumptions, including those with irregular features forming a synergy with the epistemic non-reductionism postulate.
\textbf{Deep Models:} The number of layers in a feed forward network, or the circumference of a Recurrent Neural Network (RNN) must be greater than 3, meaning multiple hidden layers in a multi-layer feed forward network, or an RNN with complex topology. Model depth avoids much of the criticism in Minsky and Papert's critical book on neural networks that showed perceptrons cannot learn concave discriminants \cite{minsky69perceptrons}, and its later editions that extend the criticism to multi-layer models. In today's ANN applications we observe all manners of intricate discrimination models were successfully learnt, however shallow networks will still not avoid Minsky's observations. A complexity analysis also supports that increasing depth can result in asymptotically smaller networks for the same function representation \cite{DBLP:conf/colt/Telgarsky16}, implying that deep models are fundamentally more efficient.
\textbf{Hierarchy and Locality:} A distinguishing feature of deep learning is that it contains local pattern recognition networks and a hierarchy of these pattern recognition circuits that affixes the local and global views. Thus, a sequence of convolutional and pooling layers have been a staple of image processing applications in deep learning as the convolutional layer is basically a set of texture recognition patches, and downsampling via max-pooling gives us a dimensionality reduction and the ability to hierarchically combine pattern recognizers efficiently. This organization was inspired by 2d image processing in the visual cortex, however many domains can benefit from the same organizational principle since they apply to any sensory array. The principle is also valid for domains that are not directly sensory arrays, but maintain a similar topological relation. The principle also has great synergy with the depth principle because the network tries to capture perceptually salient features and avoids learning irrelevant patterns making it possible to increase network depth which avoids Minskyan objections even more effectively.
\textbf{Gradient Descent:} Perhaps the most common feature of deep learning is that a variation of back propagation or gradient descent is used to train the model. This is required since any other way to train the large networks in deep learning research would be infeasible. Other methods such as variational learning and MCMC tree search have been applied in deep learning research, however this principle has remained fairly constant as it is necessitated by other principles above, which may result in billions of real valued parameters to be trained.
\textbf{Dataflow models \& SIMD acceleration} Since the number of parameters to be trained is large, exploiting data-parallelism through SIMD-based accelerators such as GPU's, and later executing data-flow representations on FPGA's have proven to be an essential factor for deep learning research. This property of deep learning corresponds to the “massive parallelism” property of the brain.

 \vspace*{-0.4cm}

\section{Shortcomings and Extensions}

 \vspace*{-0.2cm}

Although deep learning has generated phenomenal results, it also has some shortcomings that are being worked on. The most common limitation is that a typical deep learning architecture requires on the order of 10,000 or more examples. Some of the largest experiments have used millions of examples, therefore this was simply not an issue that was focused on. It may well be the case that this is a fundamental shortcoming of deep learning, however, researchers have tried solutions such as using stochastic gradient over the entire set of samples, as a usual statistical approach would necessitate, instead of running BP in epochs, which imitates the brain's online learning capability. Another common problem is that most deep learning uses supervised learning, which presents a problem in terms of constructing many labeled/annotated examples for every new problem. Autoencoder \cite{Hinton:94} is an unsupervised learning model, and it has many variations and applications in deep learning, however, most applications still require a good deal of hand crafted data. A strange problem persists in deep learning systems, which makes them easy to fool in ways that are not intuitive to humans, such as a simple perturbation causing a misclassification, an intuitively unrelated artificial image recognized as a natural image, or a specially crafted patch on an unrelated image causing a misclassification. These might either be symptoms of fundamental limitations, or they might be ameliorated with better deep learning models. We observe that these issues look much like overfitting, i.e., poor generalization performance.


When we contrast the general AI postulates and deep learning postulates, we see some interesting overlap and also some areas where deep learning requires a good deal of development.  
A deep learning system has one sort of completeness that stems from the universal approximation theorem, and dataflow models can be augmented with arbitrary computational units such as the Neural Turing Machine model \cite{ntm}, and the later Differentiable Neural Computer model \cite{dnc} that augments neural networks with external memory. Program class extensions of this sort may be an integral part of next-generation deep learning. Recent proposals for non-Euclidian embedding of data also enhance generality of deep learning models \cite{geometric-dl}. 

It is possible to design deep architectures for rigorous stochastic models, which is an important extension to deep learning that will increase robustness. 

Typically, deep learning lacks a principle of induction, but at the same time a stochastic model of induction is implicit in deep learning as the information bottleneck analysis of deep learning shows \cite{DBLP:conf/itw/TishbyZ15}, where we can view deep learning as a lossy compression scheme that forgets unnecessary information. Such theories will lead to better generalization performance. \cite{generalization-dl} applies random matrix theory to generalization in deep learning, and introduces a new regularization method for improving generalization.

Progressive deep learning architectures add layers as necessary, substantiating an important analogy to SVM's function class iteration \cite{rusu-progressive-2016}. Much richer forms of induction may be beneficial for improving a deep learning network's generalization power. The training procedure in deep learning is efficient but only locally optimal, in the future a combination of neuro-evolution and gradient descent may outperform gradient descent and approximate universal induction better. Evolution has already been applied to automated design of deep networks \cite{codeepneat,deepneuroevo}. Neuro-evolution has been shown to be effective in game playing \cite{DBLP:journals/tciaig/RisiT17} and other tasks that are difficult for deep learning, and therefore it might displace deep learning methodology altogether in the future. 

Deep learning architectures gained memory capability with the LSTM unit, and similarly designed memory cells, however, long-term memory across tasks remains problematic. A good realization of algorithmic memory in deep learning is Neural Task Programming (NTP) \cite{ntp} which achieves an indexical algorithmic memory based on LSTM and the ability to hierarchically decompose skills which has been successfully applied to robotics tasks. Progress in the direction of NTP is likely to be a major improvement for deep learning, since without cumulative and hierarchical learning intelligence is highly restricted. 

Recently, progress has been made in the matter of modularity with Hinton's update of Capsule Networks, that models the cortical architecture for visual tasks \cite{DBLP:conf/nips/SabourFH17}. Capsule Networks adds dynamic routing between visual processing modules with affine transformations, enhancing invariance and defines neural modules as capsules that may be arranged like neurons. Capsules correspond to visual entities in the model, therefore capsules that recognize a face decompose into eyes, a nose, lips, and so forth. The step from monolithic to modular deep learning is as powerful as the step from shallow to deep networks, hence this line of research is a significant extension of deep learning. A similar line of research is advanced by Vicarious, which propose a recursive neural architecture that exploits lateral connections accounting for distinct feature sets such as contour and surface, and the hierarchical representation of entities like in Capsule Networks \cite{numenta-captcha}; their system can reportedly break CAPTCHA's. Hawkins proposes a new cortex architecture that introduces pyramidal neurons, active dendrites, and multiple integration sites, identifying cortical computations for hierarchical sequence memory, and it intriguingly involves dendritic computation \cite{hawkins-2016}. Capsule Networks might be enhanced to provide a similar dendritic model eventually, or capsule-like speciation might be ported to Hawkins's model. 

Cognitive architectures built on symbolic concepts may not be readily applicable to deep learning, however, modeling the functional anatomy of the brain creates much needed synergy with neural networks. For instance, in Deep Mind's I2A model \cite{i2a}, we see a direction towards capturing more brain function in the form of imagining future states, while PathNet presents a modular, reflective learning system that can recombine network modules by evolving paths over the network \cite{pathnet}. Both neural architectures exhibit progress towards a more complete cognitive neural architecture. Another recent direction is the relational networks that model reasoning \cite{DBLP:journals/corr/SantoroRBMPBL17}. Conceivably, neural models of fundamental cognitive functions may be developed with a similar methodology, and bound in a connectionist agent architecture. Likewise, the active inference agent of \cite{friston-deeptemporal} with deep temporal models captures the essentials of functional anatomy based on hierarchical probabilistic models, and even gives us a fully unsupervised agent model that is quite intriguing from a scientific perspective.

 \vspace*{-0.4cm}

\section{Discussion and Future Research}

Despite recent criticism raised against deep learning \cite{marcus-dl-critical}, almost all of the postulates of general AI we have outlined seem achievable, however, with major improvements over existing systems. While it is entirely possible for a traditional symbolic-oriented system to achieve the same performance, the advantages of deep learning approach cannot be neglected, and the possible extensions to deep learning discussed may also ameliorate the common shortcomings we summarized. Another combination that might work is the combination of the symbolic AI approach with deep learning. In some circles, researchers pursue a mathematical AI unification approach (like AIXI approximations), however, the merits of such an approach are yet to be proven experimentally over deep learning. It seems prudent to at least try to integrate deep learning faithfully in existing AI architectures, or for new architectures, attempt to construct them solely on a neural architecture. In the future, we expect a convergence of more powerful training methods and deep architectures, taking us to a more model-free learning system, and more capable, modular neural agent architectures inspired by neuroscience.  
 \vspace*{-0.4cm}
\tiny{
\bibliographystyle{splncs04} \bibliography{agi,deep,ann,neuro}
}
\end{document}